\documentclass{article}
\usepackage{spconf,amsmath,graphicx}

\usepackage{enumitem}
\setlist{nosep, leftmargin=14pt}

\usepackage[symbol]{footmisc}
\usepackage{graphicx}
\usepackage{svg}
\usepackage{float}
\usepackage{multirow}
\usepackage{booktabs}
\usepackage[
  bookmarks=false,
  pdfpagelabels=false,
  hyperfootnotes=false,
  hyperindex=false,
  pageanchor=false,
  citecolor=black,
  colorlinks,
]{hyperref}
\usepackage[frozencache,cachedir=.]{minted}
\usepackage{amssymb}
\usepackage{pifont}
\newcommand{\cmark}{\ding{51}}%
\newcommand{\xmark}{\ding{55}}%

\usepackage{tcolorbox}
\usepackage{alphabeta}
\usepackage{fancyvrb}
\usepackage{fvextra}
\usepackage{colortbl}
\colorlet{tableheadcolor}{gray!75}
\colorlet{tablerowcolor}{gray!10}
\newcommand{\rowcol}{\rowcolor{tablerowcolor}} %

\newcommand{\emoji}[1]{\includegraphics[width=1em]{#1}}

\newcommand{\brains}{\emoji{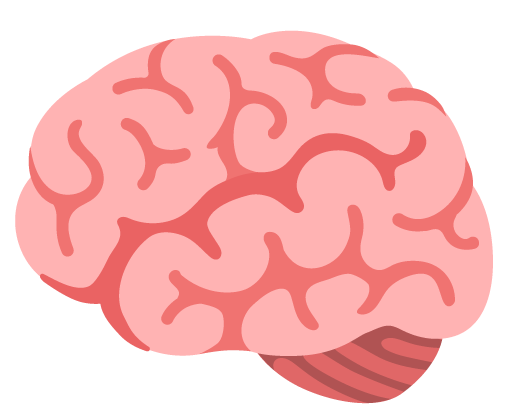}BRAINS-45K}

\usepackage{todonotes}

\title{Revisiting CLIP: Efficient alignment of 3D MRI and Tabular Data using Domain-Specific Foundation Models}
\name{Jakob Krogh Petersen\textsuperscript{*}, Valdemar Licht\textsuperscript{*}, Mads Nielsen, Asbjørn Munk}%
\address{
Pioneer Center for AI, Denmark \\
Department of Computer Science, University of Copenhagen, Denmark
\thanks{* Equal contribution}
}
\begin{document}
%
\maketitle

\begin{abstract}
Multi-modal models require aligned, shared embedding spaces. However, common CLIP-based approaches need large amounts of samples and do not natively support 3D or tabular data, both of which are crucial in the medical domain. To address these issues, we revisit CLIP-style alignment by training a \textit{domain-specific} 3D foundation model as an image encoder and demonstrate that modality alignment is feasible with only 62 MRI scans. Our approach is enabled by a simple embedding accumulation strategy required for training in 3D, which scales the amount of negative pairs across batches in order to stabilize training. We perform a thorough evaluation of various design choices, including the choice of backbone and loss functions, and evaluate the proposed methodology on zero-shot classification and image-retrieval tasks. While zero-shot image-retrieval remains challenging, zero-shot classification results demonstrate that the proposed approach can meaningfully align the representations of 3D MRI with tabular data. Code and model checkpoints are available \href{https://github.com/jakekrogh/3d-clip-for-brain-mri}{here}.
\end{abstract}


\section{Introduction}
In the pursuit of general purpose multi-modal medical models, it has become of great interest to learn aligned, shared embedding spaces. The dominant method for learning joint-embedding spaces is the contrastive language-image pre-training (CLIP) objective \cite{CLIPpaper}. By leveraging 100 million captioned images, CLIP aligns textual and image representations by maximizing the cosine similarity between positive pairs while minimizing the negative pair similarity. Due to the inherent semantic nature of natural language, CLIP shows that their embeddings efficiently transfer to downstream tasks and notably achieves high performance in a long range of zero-shot classification tasks.

While the performance of CLIP is impressive, ConVIRT~\cite{convirt} shows that due to the very high inter-class similarity present in medical imaging tasks, using medical, domain-specific contrastive training by far outperforms training on dataset of natural images. ConVIRT obtains a joint-embedding space by training on a dataset of 265\textit{k} 2D radiography images paired with textual reports and show that the learned embeddings are meaningfully aligned by doing zero-shot classification and image-retrieval.

Very little work has focused on aligning textual representations and 3D medical data, such as MRI. BrainCLIP~\cite{brainclip} uses frozen CLIP encoders for brain decoding, aligning the representations of fMRIs obtained while showing subjects images from ImageNet, with the representations of those images \textit{and} their textual representations. However, BrainCLIP does not natively train CLIP on the fMRIs in 3D, but instead uses a VAE to embed fMRI into an embedding space of frozen CLIP encoders. A gap exists in the literature on how to train CLIP models on 3D medical data directly.


\begin{figure}
    \centering
    \includegraphics[width=\linewidth]{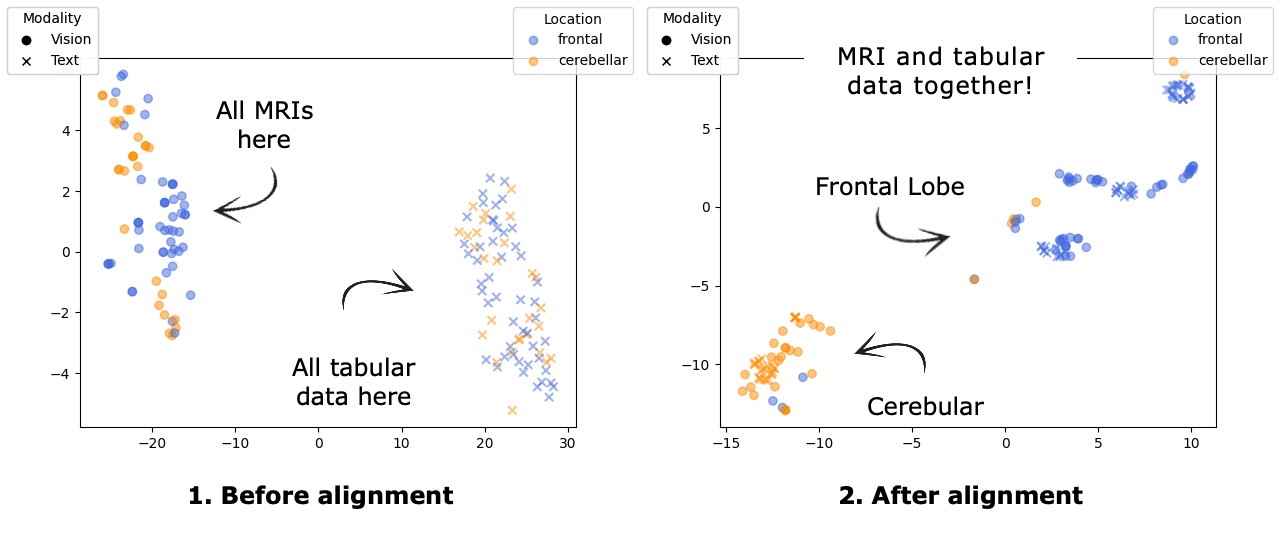}
    \caption{t-SNE \cite{tsne} visualizations of MRI scans and tabular data in the joint embedding space before and after CLIP training. Input data is the test set, with augmentations applied to increase sample size. Best viewed in color.}
    \label{fig:visual_abstract}
\end{figure}

In this paper we address this gap by examining the challenging task of learning a shared embedding space between tabular data and 3D brain MRIs. The contributions of this paper are summarized to:
\begin{itemize}
    \item We introduce, to the best of our knowledge, the first application of CLIP-training to brain MRI in native 3D.
    \item We train a \textit{domain-specific} foundation model image encoder and show that it is instrumental in enabling cross-modality alignment, in particular when using small datasets.
    \item We show that when training with small batch sizes in 3D, the contrastive loss can be stabilized by accumulating embeddings across batches, scaling the amount of negative pairs in the CLIP loss.
    \item We published both code and model weights, which can be accessed \href{https://github.com/jakekrogh/3d-clip-for-brain-mri}{here}.
\end{itemize}

\section{Method}

Starting from the standard CLIP-setup \cite{CLIPpaper}, we revisit four core design decisions in CLIP-style training: Encoding 3D MRI (§\ref{sec:method:image_encoder}), encoding the tabular data (§\ref{sec:method:text_encoder}), calculating the CLIP-loss across batches (§\ref{sec:method:batch_accumulation}) and obtaining stable training caused by the 3D input (§\ref{sec:method:tricks}). In the following, we directly ablate the result of each design decision. The experimental setup is described in §\ref{sec:experimental_setup}.

\subsection{Foundation Model Encoders}
\label{sec:method:encoders}
Central to our method is the observation that when working with datasets of very limited size, we cannot both learn a meaningful representation for each modality \textit{and} align these representations simultaneously. Instead our approach is to first learn modality-specific encoders and \textit{then} align them to a coherent embedding space.

\subsubsection{Encoding 3D MRI using domain-specific pretraining}
\label{sec:method:image_encoder}

 To obtain such an encoder for 3D brain MRI, we first perform large-scale, foundation model pretraining on a publicly available MRI dataset using the AMAES framework and the \brains~ dataset \cite{amaes}. AMAES is a framework for foundation model pretraining based on a self-supervised Masked Autoencoder strategy. \brains~ contains 44,756 brain MRI volumes from various public sources. The dataset is the largest publicly available pretraining dataset for brain MRI and contains a diverse set of sequence types. We train for 100 epochs using the default AMAES hyperparameters. We pretrain three different encoder-backbones:
\begin{itemize}
    \item \textbf{Swin-T} \cite{swin}. The 3D Swin-Transformer encoder is extracted from a pretrained SwinUNETR \cite{swinunetr}.
    \item \textbf{MedNeXt} \cite{roy2023mednext}. We use the MedNeXt-S with kernel size 3.
    \item \textbf{Resnet} \cite{resnet,cnn}. The 3D Resnet encoder is extracted from a ResUNet architecture \cite{isensee2024nnunet3d}.
\end{itemize}
\vspace{0.5em}
\noindent During CLIP-training, we increase the robustness of the embedding by applying a suite of augmentations from \cite{yucca, sebaug, amaes}. The augmentations include elastic deformations, rotation, scaling, and various noise augmentations.

\subsubsection{Encoding tabular data}
\label{sec:method:text_encoder}
Based on the observation mentioned in §\ref{sec:method:encoders}, we encode the tabular data using a BERT model \cite{bert}. We use a pretrained BERT-Base model with 110M parameters made available by the authors. The tabular data is processed by first transforming it into a natural language sentence using simple template functions. For instance, the tabular data with attributes 
\begin{verbatim}
    age_at_diagnosis: 57
    gender: female
    lesions: ['frontal', 'occipital']
\end{verbatim}
is transformed into the sentence: \\
\textit{\indent "The age of the subject is 57. The gender of the patient is female. A tumor has been identified in the frontal area of the brain. Additionally, a lesion is present in the occipital area."}

This sentence is then encoded using the BERT-encoder during alignment. The approach allows the method to be easily extended to a long range of medical data including free-text (e.g. notes from medical professionals) and semi-structured data (e.g. radiology reports).

\begin{table}[t]
\caption{\textbf{Encoder ablation}: Training is not possible without using pretrained encoders. Without text-encoder the training never becomes stable, while pretraining the vision encoder boosts performance. Metric is AUC of zero-shot classification of brain lesion location. The image encoder is Swin-T.}
\label{tab:encoders}
\centering
\resizebox{0.4\textwidth}{!}{
\begin{tabular}{cccc}
\multicolumn{2}{c}{Foundation model}                                 & \multicolumn{2}{c}{AUC}                             \\
image encoder                                & text encoder          & Cerebellar               & Average                  \\ \hline
\xmark    & \xmark   & $0.50 \pm 0.0$ & $0.50 \pm 0.0$       \\
\cmark  &   \xmark   & $0.50 \pm 0.0$ & $0.50 \pm 0.0$ \\
\xmark    & \cmark   & $0.79 \pm 0.02$ & $0.67 \pm 0.02$  \\
\rowcol \cmark    & \cmark & $\mathbf{0.87 \pm 0.1}$ & $\mathbf{0.72 \pm 0.01}$ \\
\hline
\end{tabular}
}
\end{table}
\begin{table}[t]
\caption{\textbf{Embedding accumulation ablation:} We find that larger batch sizes is important to stabilize training and improves performance. The base batch size is $8$, which is scaled by the accumulation frequency $N$. Metric is mean AUC of zero-shot classification of brain lesion location.}
\label{tab:batch}
\centering
\resizebox{0.4\textwidth}{!}{
\begin{tabular}{lccc}
                              & \multicolumn{3}{c}{Accumulation Frequency} \\
Encoder                       & None               & N = 8             & N = 16      \\ \hline
MedNeXt                       & $0.64 \pm 0.01$    & $0.65 \pm 0.02$    & $0.64 \pm 0.01$\\
Resnet                        & $0.61 \pm 0.03$    & $0.71 \pm 0.01$   & $0.71 \pm 0.01$ \\
\rowcol Swin-T & $0.64 \pm 0.01$    & $0.69 \pm 0.02$  & $\mathbf{0.72 \pm 0.01}$ \\ \hline
\end{tabular}
}
\end{table}


\subsubsection{Effect of using foundation model encoders}

To investigate the effect of using the above foundation model encoders, we perform CLIP-training with encoders initialized with foundation model weights and random weights. The result is given in Table \ref{tab:encoders}. We expect the model to perform very poorly in particular when there is no language model, since the tabular data is encoded as unstructured free-text. We find that the 3D MRI encoder provide a $+0.05$ boost in average AUC.

\subsection{Computing the CLIP-loss across batches}
\label{sec:method:batch_accumulation}
The large size of 3D volume representations significantly reduces the maximum feasible batch size during training. To alleviate this constraint, we construct a simple embedding accumulation method, which allows the scaling of negative samples when computing the loss. For accumulation frequency $N$ and batch size $B$ we compute embeddings for all $N$ batches, and use these embeddings to scale the negative examples across batches. The exact computation is given in Figure \ref{fig:batch_accum}. Computing the loss and backpropagating at each iteration of $j$ results in the effective computation of the CLIP loss with a batch size $N \cdot B$ while only tracking the gradient batches with size $B$. Note that the optimization is computed when all gradients are accumulated. Implementation is adapted from \cite{openCLIP}. The effect of computing the CLIP-loss across batches is given in Table \ref{tab:batch}.
\begin{figure}[t!]
    \centering
    \small
        \begin{minted}{python}
def accumulated_loss(N, X_t, X_i):
    # N: accumulation frequency.
    # X_t, X_i: lists containing N batches.
    
    Y_t, Y_i = [], []
    for i = 0 to N and without_gradients():
        y_t = text_encoder(X_t[i]).detach()
        y_i = image_encoder(X_i[i]).detach()
        Y_t.append(y_t), Y_i.append(y_i)

    loss = 0
    for j = 0 to N: 
        y_t = text_encoder(X_t[j])
        y_i = image_encoder(X_i[j])
        C_i = concat(Y_i[:j], y_i, Y_i[j+1:])
        C_t = concat(Y_t[:j], y_t, Y_t[j+1:])
        loss += loss(C_i, C_t)
    loss.backwards()
    optimizer.step()
        \end{minted}
    \caption{Pseudo code showing how accumulated loss over $N$ batches is calculated. The embedding accumulation strategy computes embeddings for all input in $N$ batches, and for each batch uses negative examples from all batches to compute the loss and update the model.}
    \label{fig:batch_accum}
\end{figure}


\subsection{Obtaining stable training in 3D}
\label{sec:method:tricks}
Training in 3D is challenging. In our method, models are trained in native 3D, by extracting sub-patches from each volume of size $64^3$ for $50.000$ steps using the AdamW optimizer and a cosine annealing learning rate scheduler. Additional to a larger batch size as described in §\ref{sec:method:batch_accumulation}, we found three things to be important in order to obtain stable training and avoiding collapse: 1. Using a warm-up of the learning rate for $2000$ steps using a linear increasing scheduler from $0$ to the starting learning rate. 2. The vision encoder needed to have an order of magnitude higher starting learning rate than the text encoder. 3. The temperature-parameter in the CLIP loss needed to be carefully tuned.

\section{Experimental setup}
\label{sec:experimental_setup}
We validate the effectiveness of our embedding alignment on two challenging tasks: zero-shot classification and image-retrieval of tumor location from brain MRI using the Brain-TR-GammaKnife dataset \cite{gammaknife}. The dataset is created to study the recurrence of brain tumors after Gamma-Knife Radiotherapy. We use this dataset since it contains dense annotations on the lesion level with auxiliary tabular information about each scan, such as age, gender, and brain tumor location. The following section describes the setup used to obtain our results.

\subsection{Data}
Data is T1 MPRAGE with Gadolinium contrast MRI volumes acquired using a 1.5 T Siemens Magnetom scanner with 1mm isotropic resolution. The dataset contains MRI volumes from $47$ different brain cancer patients with $21$ males and $26$ females ranging in age from $23$ to $77$ years who underwent Gamma Knife radiation therapy. Most patients had $1$–$2$ treatment courses, but the dataset includes outliers with up to eight treatment courses, leading to a total of $77$ MRI scans. A total of $244$ lesions were collected and annotated during treatment planning. We split the data into training and test sets using a $80/20$ split on the scan level. For validation we use 5-fold cross validation on the training set. We use the validation split for hyperparameter tuning. The training dataset contains $62$ MRI scans.


\textbf{Region Labels}. To perform our evaluation, we categorize the annotated lesions in the dataset into 5 distinct labels. We base the first four labels on the largest and most discriminatory regions in the cerebral cortex: frontal lobe, parietal lobe, temporal lobe, and occipital lobe. Finally, we include the cerebellar, as the region is well represented in the dataset. Based on this categorization, we define a subset of the dataset containing 222 lesions.


\textbf{Preprocessing}. All MRI volumes in the dataset are preprocessed using the Yucca framework \cite{yucca}. The scans are downsampled to a resolution of $192^3$ to allow for smaller patch sizes that still contain anatomical information. The volumes are bias-field-corrected and transformed into the RAS+ orientation. Each volume is independently normalized using z-normalization.

\begin{table*}[t!]
\caption{\textbf{Results for zero-shot classification} of lesion location for all models averaged over five folds. Metric is average AUC with standard error of the mean.}
\label{tab:classification}
\centering
\begin{tabular}{lccccccc}
Encoder   & Params \textit{(M)} & \multicolumn{1}{c}{Temporal} & \multicolumn{1}{c}{Frontal} & \multicolumn{1}{c}{Occipital} & \multicolumn{1}{c}{Parietal} & \multicolumn{1}{c}{Cerebellar} & Avg                      \\ \hline
MedNeXt & 4    & $0.48 \pm 0.07$ & $0.77 \pm 0.03$  & $\mathbf{0.73 \pm 0.05}$  & $0.51 \pm 0.06$  & $0.76 \pm 0.03$   & $0.65 \pm 0.02$          \\
Resnet          & 57 & $\mathbf{0.66 \pm 0.05}$ & $0.79 \pm 0.02$    & $0.72 \pm 0.04$      & $\mathbf{0.54 \pm 0.02}$     & $\mathbf{0.88 \pm 0.03}$                & $0.71 \pm 0.01$          \\
\rowcol Swin-T  & 8 & $0.62 \pm 0.05$ & $\mathbf{0.85 \pm 0.01}$ & $\mathbf{0.73 \pm 0.03}$  & $0.48 \pm 0.06$ & $0.87 \pm 0.01$  & $\mathbf{0.72 \pm 0.01}$ \\
\hline
\end{tabular}
\end{table*}

\begin{table}[t!]
\caption{\textbf{Results for zero-shot image retrieval} for all models averaged over five folds with standard error of the mean.}
\label{tab:ir}
\centering
\begin{tabular}{lcccc}
Encoder   & \multicolumn{1}{c}{MMR} & \multicolumn{1}{c}{mAP} & \multicolumn{1}{c}{Accuracy} \\ \hline
\textit{Random} & $0.21 \pm 0.02$                    & $0.21 \pm 0.001$                    & $0.19 \pm 0.001$                         \\
MedNeXt & $0.25  \pm 0.01$                    & $0.30  \pm 0.02$                    & $0.27  \pm 0.02$                         \\
Resnet  & $0.18  \pm 0.02$                    & $0.19  \pm 0.01$                    & $0.23  \pm 0.01$                         \\
\rowcol Swin-T  &  $\mathbf{0.40 \pm 0.03}$        & $\mathbf{0.34 \pm 0.02}$           & $\mathbf{0.35 \pm 0.02}$ \\ \hline
\end{tabular}
\end{table}

\subsection{Hyperparameters and experimental details}
We do careful hyperparameter tuning on the validation splits. In particular, we find that the optimal batch size for MedNeXt and Resnet models were 64 (accumulation frequency $N=8$) and for the Swin-T the optimal batch size was 128 ($N=16$). Further, we used initial temperature $\tau = 1.351$ in the CLIP-loss and an embedding space with 512 dimensions. All experiments were run on single Titan RTX GPUs with 24 GB VRAM.

\subsection{Evaluation Tasks}

To evaluate the alignment of the shared embeddings, we seek to explore the relation between semantics in one modality from the other. By using zero-shot classification and image-retrieval, we evaluate the alignment by going from one modality to the other. For zero-shot classification for a given scan, we obtain an image embedding, and compare the cosine similarity to embeddings of sentences describing each region. For zero-shot image-retrieval we go the other way. By embedding a sentence, we can rank all scans in the test set by their euclidean distance to the sentence embedding. For both, we embed regions using sentences of the form:
\begin{center}
    \textit{There is a lesion in the [region] section}
\end{center}
For all results, we report averages over five folds on the test dataset along with standard error of the mean.

\section{Results \& Discussion}


\noindent \textbf{Zero-shot classification}. We measure the area under the receiving operating characteristic curve (AUC). Owing to the multiclass nature of the problem, we employ the one-versus-rest (OvR) strategy, where we treat each sample as a binary classification task. Thus, we measure the probability of the model predicting the ground truth label versus all other possible labels and report the average AUC over this heuristic. Results are given in Table \ref{tab:classification}. Our results show that all models beat random, while Swin-T encoder yields the highest AUC of all models and is closely followed by the ResNet model. For the labels \texttt{frontal} and \texttt{cerebellar}, the we obtain an AUC of over $0.8$, while on labels \texttt{parietal} we obtain an AUC which is close to random. We note that the problem is significantly complicated by the fact that each patient typically has multiple tumors. \\


\noindent \textbf{Zero-shot image retrieval}. We measure both accuracy, mean reciprocal rank (MMR) and mean average precision (mAP) calculated for each label and then averaged. Results are given in Table \ref{tab:ir}. Both Swin-T and the MedNeXt models beat random, while interestingly Resnet does not.

\subsection{Discussion}

The results on the difficult zero-shot tasks, highlight that alignment of image and textual embedding spaces, is possible given pretrained domain-specific image encoders. In particular, the quality of the joint-embedding space when using a Swin-Transformer image encoder meaningfully beat random. Interestingly, the MedNext encoder outperforms the Resnet on image retrieval, while the opposite is true on classification. This highlights, that robustness of the joint-embedding space is still an issue when using very small datasets. \\

\noindent\textbf{Limitations and future work}. This study is conducted using a single dataset and the evaluation does not investigate the effectiveness of our methodology on data outside the dataset distribution. Due to the limited label space of the dataset, an evaluation of the model's zero-shot performance on a larger set of tasks is missing. Additionally, we did not have the resources to search for optimal hyperparameters for each batch size. This could significantly impact performance for models with higher batch sizes. This study opens up for future work in multiple directions: scaling the dataset size, working with free-text image captions and exploring the models performance in a \textit{few-shot} setting.

%

\section{Conclusion}
In this study, we explore CLIP-alignment using a small domain-specific dataset using domain-specific foundation model encoders. We show that tabular data transformed into natural language can supervise 3D vision tasks and that a joint embedding space can yield meaningful results on non-trivial downstream tasks in a zero-shot setting. Enabled by a embedding accumulation strategy, we scale the CLIP-loss across batches and demonstrate how it improves the stability and performance of different vision encoders. Our zero-shot classification results show that we can obtain a reasonable alignment using very few data points. 



\section{Compliance with ethical standards}
This research study was conducted retrospectively using human subject data made available in open access. Ethical approval was not required as confirmed by the license attached with the open access data.

\section{Acknowledgements}
Thanks to Jakob Ambsdorf for his comments on the manuscript. This work was supported by Danish Data Science Academy, which is funded by the Novo Nordisk Foundation (grant number NNF21SA0069429) and Villum Fonden (grant number 40516), and Pioneer Centre for AI, Danish National Research Foundation, grant number P1.

\bibliographystyle{IEEEbib}
\bibliography{strings,refs}

\end{document}